\documentclass[letterpaper, 10 pt, conference]{ieeeconf}  

\IEEEoverridecommandlockouts
\usepackage{courier}
\usepackage{graphicx,subfigure}
\usepackage{epstopdf}
\usepackage{tikz}
\usepackage{amsmath}
\usepackage{multirow}
\usepackage{amsfonts}
\usepackage{float}
\usepackage{amssymb}
\usepackage{graphicx}
\usepackage[linesnumbered,ruled,vlined]{algorithm2e}
\usepackage{mathtools}
\usepackage{pifont}
\usepackage{cite}
\usepackage{xcolor}

\newcommand{\xmark}{\ding{55}}
\usepackage{hyperref}
\hypersetup{colorlinks=true}
\usepackage{booktabs}
\usepackage{etoolbox}
\makeatletter
\patchcmd{\@makecaption}
{\scshape}
{}
{}
{}
\makeatother

\title{\LARGE \bf
	PL-CVIO: Point-Line Cooperative Visual-Inertial Odometry
}

\author{Yanyu Zhang, Pengxiang Zhu, and Wei Ren
\thanks{This work was supported by National Science Foundation under Grant CMMI-2027139.}
\thanks{Y. Zhang, P. Zhu, and W. Ren are with the Department of Electrical and Computer Engineering, University of California, Riverside, CA, 92521, USA. Email: \{yzhan831, pzhu008\}@ucr.edu, ren@ee.ucr.edu}%
}

\begin{document}

\maketitle
\thispagestyle{empty}
\pagestyle{empty}

\begin{abstract}
	Low-feature environments are one of the main Achilles' heels of geometric computer vision (CV) algorithms. In most human-built scenes often with low features, lines can be considered complements to points. In this paper, we present a multi-robot cooperative visual-inertial navigation system (VINS) using both point and line features. By utilizing the \textit{covariance intersection} (CI) update within the multi-state constraint Kalman filter (MSCKF) framework, each robot exploits not only its own point and line measurements, but also constraints of common point and common line features observed by its neighbors. The line features are parameterized and updated by utilizing the \textit{Closest Point} representation. The proposed algorithm is validated extensively in both Monte-Carlo simulations and a real-world dataset. The results show that the point-line cooperative visual-inertial odometry (PL-CVIO) outperforms the independent MSCKF and our previous work CVIO in both low-feature and rich-feature environments.
\end{abstract}

\section{INTRODUCTION AND RELATED WORK}
Simultaneous localization and mapping (SLAM) has received considerable attention in the past few decades and has already been the core technology in many robotics and computer vision applications, such as augmented/virtual reality, autonomous driving, and robot navigation. In GPS-denied environments, visual-inertial navigation systems (VINS) and related algorithms \cite{Qin1, Li1, Wu} have received considerable popularity through utilizing low-cost and lightweight onboard cameras and inertial measurement units (IMUs). However, multiple robots have the ability to accomplish tasks more efficiently and achieve higher accuracy than a single robot \cite{Huang}. Therefore, a key question for a multi-robot group is how to best utilize the environment information and other robots' information.

In human-made scenarios, lines can be considered good complements to points, especially in low-feature environments where only a few point features can be extracted. There are two main categories of methods for processing points and lines in VINS: \textit{indirect} (feature-based) and \textit{direct} methods. In particular, the indirect methods pre-process image flows by extracting feature descriptors and matching them along a sequence \cite{Qin1, Leutenegger, He, Mourikis}. The indirect methods optimize the system by minimizing the geometric error. The direct methods skip the feature extraction step and optimize the photometric error using row pixels directly \cite{Engel1,Engel2,Zhou}. The direct methods are highly efficient but need to assume brightness constancy (ignoring exposure changes), while the exposure varies heavily in the real-world environment.

\begin{figure}[t]
	\centering
	\subfigure[ ]{
		\includegraphics[width=7cm, height=4cm]{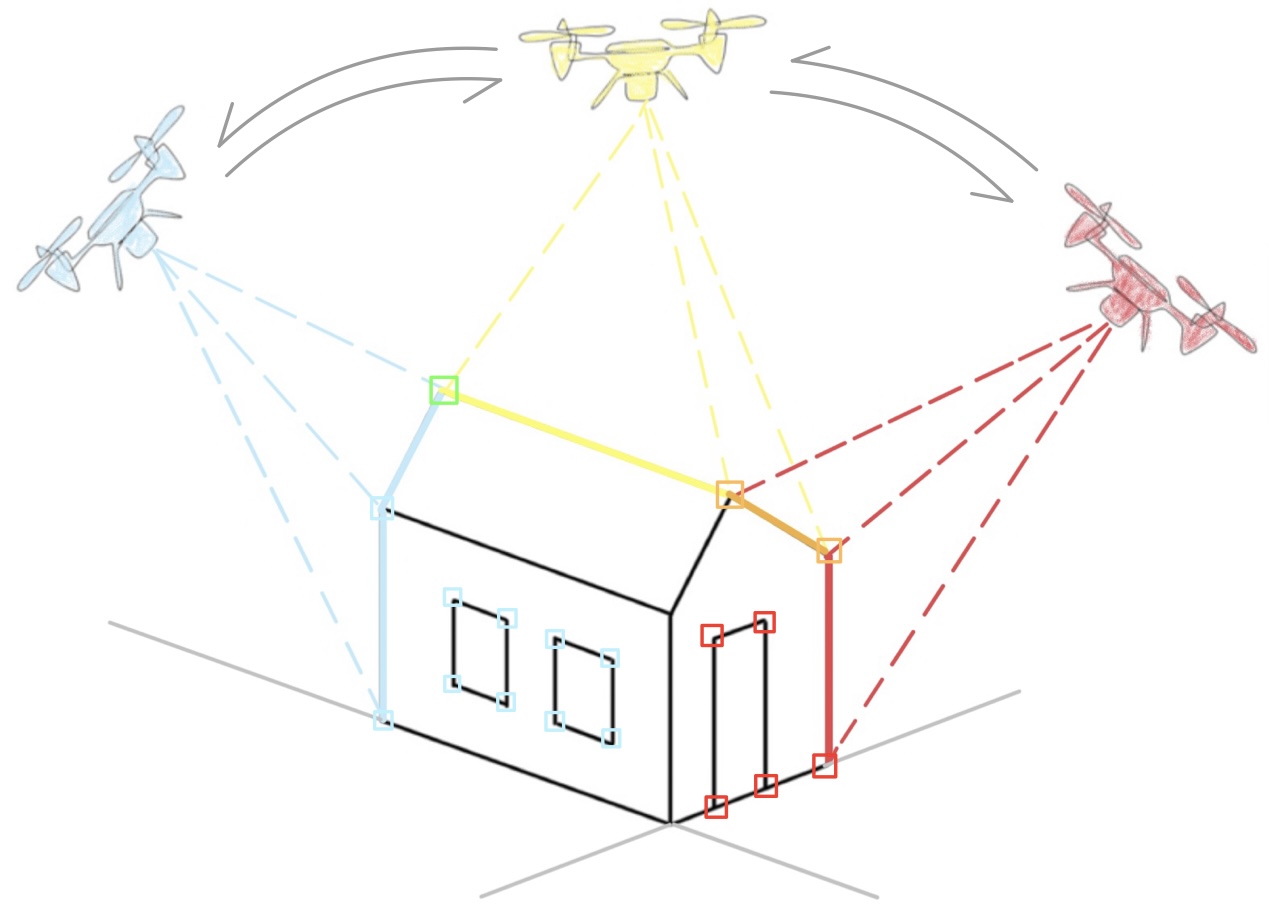}}
	\subfigure[ ]{
		\includegraphics[width=2.7cm, height=2.7cm]{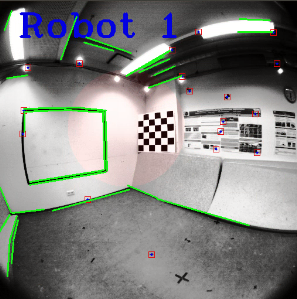}
		\includegraphics[width=2.7cm, height=2.7cm]{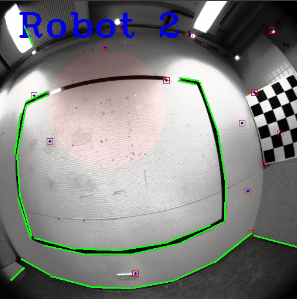}
		\includegraphics[width=2.7cm, height=2.7cm]{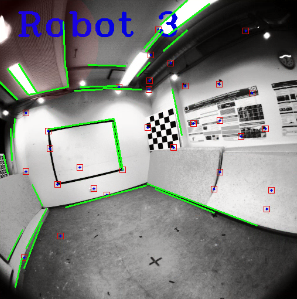}}
	\caption{{\bf (a)} Overview of the PL-CVIO. Here multiple robots observe point (square) and line (line segment) features in the same environment, neighbors communicate and share common points (green and orange squares) and common lines (orange line segment), and PL-CVIO is performed to estimate the global poses of each robot. {\bf (b)} Point and line feature detection of three different robots in the TUM dataset \cite{Sturm}. Here a green edge denotes a line extracted from the current frame, and a blue dot surrounded by a red square denotes a point extracted from the current frame.}
	\label{fig:overview}
\end{figure}

Among the previous feature-based VINS literature, the solutions can be broadly classified into two categories: filter-based methods \cite{Mourikis, Bloesch, Forster, Paul, Yu, Yang1, Yang2} and graph-based methods \cite{Qin1, Campos, He, Leutenegger, Pumarola, Mur-Artal}. One of the \textit{state-of-the-art} works of the filter-based methods is the multi-state constraint Kalman filter (MSCKF) \cite{Mourikis}, which formed a multi-constraint update by using the measurements of the same feature. A tightly coupled monocular graph-based VIO (VINS-Mono) and nonlinear optimization with robust initialization introduced in \cite{Qin1}. Besides, there are also some VINS algorithms using both point and line features. The point-line visual-inertial odometry (PL-VIO) \cite{He} is an extension of VINS-Mono, which can optimize the re-projection errors of the point and line features in a sliding window. PL-SLAM \cite{Pumarola} proposed a point-line SLAM framework based on ORB-SLAM \cite{Mur-Artal}. Line features used in Plücker representation for rolling-shutter cameras were designed in \cite{Yu}. Article \cite{Yang1} proposed two line triangulation algorithms. The analysis of three different line representations (Plücker, Quaternion, Closest Point) and the corresponding observabilities were provided in \cite{Yang2}. However, all of the above references focus on the single robot case.

One advantage of the cooperative VINS (C-VINS) is the sharing of common features from multiple robots so as to introduce more geometric constraints of the common features. In particular, each robot in the group not only observes its own measurements like in the previous literature, but also collects measurements from the multi-robot group. The robot applies an update to improve the localization performance by utilizing the common feature constraints. There exist some centralized multi-robot solutions \cite{Luna, Liu, Kang}. They usually require expensive computation and communication. Distributed algorithms offer some benefits in this regard. Recently, \cite{Benedettelli} provided a distributed point-line cooperative SLAM (C-SLAM) algorithm by adopting the M-Space representation of different kinds of features, but the consistency of the estimation cannot be guaranteed because of repeated usage of the same information in the robot group. In \cite{Karam}, each robot in the group processed its own available measurements, and fused the estimation and covariance with other robots within the communication range only at a particular time step. DOOR-SLAM \cite{Lajoie} introduced a fully distributed C-SLAM algorithm that contains a pose graph optimizer model and a data-efficient distributed SLAM frontend similar to \cite{Cieslewski}. Article \cite{Nerurkar} proposed a fully distributed algorithm using the maximum a posteriori (MAP). Our previous work CVIO \cite{Zhu} provided a fully distributed cooperative algorithm and can guarantee consistency by utilizing the covariance intersection (CI) update, but the low-feature environments were not considered.
 
In this paper, we propose a fully distributed multi-robot pose estimation algorithm using both point and line features. Each robot not only exploits its own point and line measurements, but also resorts to the cooperation with neighbors (see Fig.\ \ref{fig:overview}). Especially in low-feature environments, where robust landmarks are absent, each robot's pose can be estimated with high accuracy by fusing independent point and line features from itself and utilizing the CI update to exploit the constraints imposed by commonly observed point and line features by neighbors. The PL-CVIO algorithm is developed in the \textit{state-of-the-art} OpenVINS \cite{Geneva} system using the monocular camera-IMU architecture. Monte-Carlo simulations and real-world experiments are used to validate the performance of our PL-CVIO algorithm. In both low-feature and rich-feature environments, our algorithm is shown to achieve more accurate localization.

\section{PROBLEM FORMULATION}
The goal of the cooperative point and line visual-inertial estimator is to track the 3D pose of each robot \{$I_i$\}, for $i = 1, \cdots, n$ in the global frame \{$G$\}. Unlike the independent case, multiple robots can share common features with neighbors. In this paper, we utilize both common point and common line features to improve the localization accuracy.

\subsection{Visual-Inertial Odometry State Vector}
\label{section:vio}
In order to perform the PL-CVIO, the state vector of each robot $i$ is defined as: 
\begin{equation}
	\begin{aligned}
		\mathbf{x}_i=\left[\begin{array}{cccc}
			\mathbf{x}_{I_i}^{\top} & \mathbf{x}_{Calib_i}^{\top} & \mathbf{x}_{C_i}^{\top} &  t_{d_i}
		\end{array}\right]^{\top},
		\label{eq:state1}
	\end{aligned}
\end{equation}
where $\mathbf{x}_{I_i}$ denotes the IMU state vector, $\mathbf{x}_{Calib_i}$ denotes the rigid body tranformation between the IMU frame and camera frame, $\mathbf{x}_{C_i}$ represents the cloned IMU states, and $t_{d_i} = t_{C_i} - t_{I_i}$ denotes the time-offset between robot $i$'s camera \{$C_i$\} clock and IMU clock, which treats the IMU clock as the true time \cite{Li2, Qin2}. At any time step $k$, the state vector of each IMU can be writen as:
\begin{equation}
	\begin{aligned}
		\mathbf{x}_{I_{i, k}}=\left[\begin{array}{ccccc}
		    { }_{G}^{I_{i, k}} \bar{q}^\top & 
			{ }^{G} \mathbf{p}_{I_{i, k}}^{\top} & 
		    { }^{G} \mathbf{v}_{I_{i, k}}^{\top} & 
		    \mathbf{b}_{g_{i, k}}^{\top} & 
		    \mathbf{b}_{a_{i, k}}^{\top}
		\end{array}\right]^{\top},
		\label{eq:state2}
	\end{aligned}
\end{equation}
where ${ }_{G}^{I_{i, k}} \bar{q}$ denotes the JPL unit quaternion \cite{Trawny} representing the rotation from the global frame to the IMU frame at time step $k$. ${ }^{G} \mathbf{p}_{I_{i, k}}$ and ${ }^{G} \mathbf{v}_{I_{i, k}}$ are the IMU position and velocity in the global frame at time step $k$. $\mathbf{b}_{g_{i, k}}$ and $\mathbf{b}_{a_{i, k}}$ are the gyroscope and accelerometer biases at time step $k$. Then, the error state of the IMU is defined as:
\begin{equation}
	\begin{aligned}
		 \mathbf{\tilde{x}}_{I_{i, k}}=\left[\begin{array}{ccccc}
			\delta_{G}^{I_{i, k}} \boldsymbol{\theta}^{\top} & 
			{ }^{G} \mathbf{\tilde{p}}_{I_{i, k}}^{\top} & 
			{ }^{G} \mathbf{\tilde{v}}_{I_{i, k}}^{\top} & 
			\mathbf{\tilde{b}}_{g_{i, k}}^{\top} & 
			\mathbf{\tilde{b}}_{a_{i, k}}^{\top}
		\end{array}\right]^{\top},
		\label{eq:state3}
	\end{aligned}
\end{equation}
where the position, velocity, and bias errors utilize the standard additive error, while the quaternion error state is described by
\begin{equation}
	\begin{aligned}
		\bar{q}=\delta \bar{q} \otimes \hat{\bar{q}} \simeq\left[\frac{1}{2} \delta \boldsymbol{\theta}^{\top} \quad 1\right]^{\top} \otimes \hat{\bar{q}},
	    \label{eq:state4}
    \end{aligned}
\end{equation}
where $\hat{\left(\cdot \right)}$ denotes the estimate, and $\otimes$ is the quaternion multiplication operator.

In addition to robot $i$'s IMU state, the spatial calibration between its IMU frame and camera frame will also be estimated. In particular, the calibration state vector contains the unit quaternion rotation from the IMU frame to the camera frame ${ }_{I_i}^{C_i} \bar{q}$, and the translation from the IMU frame to the camera frame ${ }^{C_i} \mathbf{p}_{I_i}$ as:
\begin{equation}
	\begin{aligned}
	        \mathbf{x}_{Calib_{i}}=\left[\begin{array}{cc}
			{ }_{I_i}^{C_i} \bar{q}^\top & 
			{ }^{C_i} \mathbf{p}_{I_i}^{\top}
		\end{array}\right]^{\top}.
		\label{eq:state5}
	\end{aligned}
\end{equation}
Robot $i$ maintains a sliding window with $m$ cloned IMU poses at time step $k$ written as:
\begin{equation}
	\begin{aligned}
		\mathbf{x}_{C_{i, k}}=\left[{ }_{G}^{I_{i, k-1}} \bar{q}^\top \quad 
		{ }^{G} \mathbf{p}_{I_{i, k-1}}^{\top} \quad
		...\quad
        { }_{G}^{I_{i, k-m}} \bar{q}^\top \quad
        { }^{G} \mathbf{p}_{I_{i, k-m}}^{\top}
		\right]^{\top}.
		\label{eq:state6}
	\end{aligned}
\end{equation}

\subsection{Dynamic System Model}
\label{section:dynamic}
For each robot $i$, the measurement of the IMU linear acceleration ${ }^{I_i} \mathbf{a}_{m}$ and the angular velocity $	{ }^{I_i} \boldsymbol{\omega}_{m}$ are modeled as:
\begin{align}
	& { }^{I_i} \mathbf{a}_{m}={ }^{I_i} \mathbf{a}+{ }_{G}^{I_i} \mathbf{R}^{G} \mathbf{g}+\mathbf{b}_{a_i}+\mathbf{n}_{a_i}, \\
	& { }^{I_i} \boldsymbol{\omega}_{m}={ }^{I_i} \boldsymbol{\omega}+\mathbf{b}_{g_i}+\mathbf{n}_{g_i},
	\label{eq:model1}
\end{align}
where ${ }^{I_i} \mathbf{a}$ and ${ }^{I_i} \boldsymbol{\omega}$ are the true angular velocity and linear acceleration. $\mathbf{n}_{a_i}$ and $\mathbf{n}_{g_i}$ represent the continuous-time Gaussian noises that contaminate the IMU measurements. ${}^{G} \mathbf{g}$ denotes the gravity expressed in the global frame. Then, the dynamic system of each IMU can be modeled as \cite{Trawny}:
\begin{align}
	\begin{split}
		& { }_{G}^{I_i} \dot{\bar{q}}(t)=\frac{1}{2} \mathbf{\Omega}\left({ }^{I_i} \boldsymbol{\omega}(t)\right){ }_{G}^{I_i} \bar{q}(t), \quad\dot{\mathbf{b}}_{g_i}(t)=\mathbf{n}_{{w g}_i}(t), \\ 
		& { }^{G} \dot{\mathbf{v}}_{I_i}(t)={ }^{G} \mathbf{a}_i(t), \quad \dot{\mathbf{b}}_{a_i}(t)=\mathbf{n}_{{w a}_i}(t), \quad{ }^{G} \dot{\mathbf{p}}_{I_i}(t)={ }^{G} \mathbf{v}_{I_i}(t)
		\label{eq:model2}
    \end{split}
\end{align}
where ${ }^{G} \mathbf{a}_i$ is the body acceleration in the global frame. ${ }^{G} \mathbf{v}_{I_i}$, ${ }^{G} \mathbf{p}_{I_i}$ are the velocity and position of the IMU in the global frame. $\mathbf{n}_{{w g}_i}$ and $\mathbf{n}_{{w a}_i}$ denote the zero-mean Gaussian noises driving the IMU biases. $\boldsymbol{\omega} = [\omega_x \ \omega_y \ \omega_z]^{\top}$ is the rotational velocity in the IMU frame and
\begin{align}
	\nonumber
	\boldsymbol{\Omega}(\boldsymbol{\omega})=\left[\begin{array}{cc}
	-\lfloor\boldsymbol{\omega} \times\rfloor & \boldsymbol{\omega} \\
	-\boldsymbol{\omega}^{T} & 0\end{array}\right], 
    \lfloor\boldsymbol{\omega} \times\rfloor=\left[\begin{array}{ccc}
	0 & -\omega_{z} & \omega_{y} \\
	\omega_{z} & 0 & -\omega_{x} \\
	-\omega_{y} & \omega_{x} & 0
\end{array}\right].
\end{align}
After linearization, the continuous-time IMU error-state can be written as: 
\begin{equation}
	\begin{aligned}
		\dot{\tilde{\mathbf{x}}}_i(t) \simeq
			\mathbf{F}_i(t) \tilde{\mathbf{x}}_i(t) + \mathbf{G}_i(t) \mathbf{n}_i(t),
    \label{eq:model3}
    \end{aligned}
\end{equation}
where $\mathbf{F}_i(t)$ is the $15\times15$ continuous-time IMU error-state Jacobian matrix, $\mathbf{G}_i(t)$ is the $15\times12$ noise Jacobian matrix, and $\mathbf{n}_i(t) = \left[\mathbf{n}_{g_i}^{\top} \ \mathbf{n}_{wg_i}^{\top} \ \mathbf{n}_{a_i}^{\top} \ \mathbf{n}_{wa_i}^{\top}\right]^{\top}$ is the system noise with the covariance matrix $\mathbf{Q}_i$.

In order to propagate the covariance matrix from discrete-time $t_k$ to $t_{k+1}$, the state transition matrix $\boldsymbol{\Phi}_i\left(t_{k+1}, t_{k}\right)$ is computed by solving the differential equation:
\begin{equation}
	\begin{aligned}
		\dot{\boldsymbol{\Phi}_i}\left(t_{k+1}, t_{k}\right)=\mathbf{F}_i \boldsymbol{\Phi}_i\left(t_{k+1}, t_{k}\right),
        \label{eq:model4}
    \end{aligned}
\end{equation}
with the initial condition $\boldsymbol{\Phi}_i\left(t_{k}, t_{k}\right) = \mathbf{I}_{15}$. Thus, the discrete-time noise covariance can be expressed as:
\begin{equation}
	\begin{aligned}
		\mathbf{Q}_{i, k}=\int_{t_{k}}^{t_{k+1}} \boldsymbol{\Phi}_i (t_{k+1}, \tau) \mathbf{G}_i(\tau) \mathbf{Q}_i  \mathbf{G}_i^{\top}(\tau) \boldsymbol{\Phi}_i (t_{k+1}, \tau)^{\top} \mathbf{d} \tau,
		\label{eq:model5}
	\end{aligned}
\end{equation}
and the propagated covariance can be written as:
\begin{equation}
	\begin{aligned}
		\mathbf{P}_{i, k+1 \mid k}=\boldsymbol{\Phi}_i\left(t_{k+1}, t_{k}\right) \mathbf{P}_{i, k \mid k} \boldsymbol{\Phi}_i\left(t_{k+1}, t_{k}\right)^{\top}+\mathbf{Q}_{i, k}.
		\label{eq:model6}
	\end{aligned}
\end{equation}

\subsection{Point and Line Measurement Models}
In low-feature environments, lines are good complements to points. Hence we consider both point and line measurements in this paper. The point measurements of robot $i$ can be described by:
\begin{equation}
	\begin{aligned}
		{ }^{C_i} \mathbf{z}_{p}=\Pi \left({ }^{C_i} \mathbf{x}_{p}\right)+\mathbf{w}_{p_i}, \quad \Pi\left(\left[
			x \ y \ z \right]^{\top}\right)=\left[
		\frac{x}{z} \quad \frac{y}{z} \right]^{\top},
		\label{eq:pt1}
    \end{aligned}
\end{equation}
where ${ }^{C_i} \mathbf{x}_{p}$ is the 3D position of the point in the camera frame, and $\mathbf{w}_{p_i}$ denotes the corresponding measurement noise. Based on the relative transformation and time offset definition in (\ref{eq:state1}), the relationship between point feature in the global frame ${ }^{G} \mathbf{x}_{p}$ and in the camera frame ${ }^{C_i} \mathbf{x}_{p}$ can be expressed as: 
\begin{equation}
	\begin{aligned}
		{ }^{C_i} \mathbf{x}_{p}={ }_{I_i}^{C_i} \mathbf{R}_{G}^{I_i} \mathbf{R}\left(\bar t_i \right)\left({ }^{G} \mathbf{x}_{p}-{ }^{G} \mathbf{p}_{I_i}\left(\bar t_i \right)\right)+{ }^{C_i} \mathbf{p}_{I_i},
		\label{eq:pt2}
	\end{aligned}
\end{equation}
where $\bar t_i = t_i-t_{d_i}$ is the exact camera time of the relative transformation between the global frame and the IMU frame.

For a 3D line, we adopt the \textit{Closest Point} representation \cite{Yang1}, which represents the 3D line by multiplying a unit quaternion and the corresponding distance scalar from the origin to this line. Given the 3D positions of two points $\mathbf{p_{f1}}$ and $\mathbf{p_{f2}}$ on a line, the Plücker coordinate can be expressed by \cite{Zuo}:
\begin{equation}
	\begin{aligned}
		\left[\begin{array}{l}
			\mathbf{n}_{l} \\
			\mathbf{v}_{l}
		\end{array}\right]=\left[\begin{array}{l}
			\left\lfloor\mathbf{p}_{\mathbf{f} 1}\times\right\rfloor \mathbf{p}_{\mathbf{f} 2} \\
			\mathbf{p}_{\mathbf{f} 2}-\mathbf{p}_{\mathbf{f} 1}
		\end{array}\right],
		\label{eq:ln1}
	\end{aligned}
\end{equation}
where $\mathbf{n}_{l}$ denotes the normal direction of the line-plane and $\mathbf{v}_{l}$ is the line direction. Then, the \textit{Closest Point} line can be expressed as:
\begin{equation}
	\begin{aligned}
		{ }^{G}\mathbf{x}_l = d_l \bar{q_l} = \left[\mathbf{q}_l^{\top} \quad q_l \right]^{\top},
		\label{eq:ln2}
	\end{aligned}
\end{equation}
where the distance scalar can be computed as $d_{l}=\left\|\mathbf{n}_{l}\right\| /\left\|\mathbf{v}_{l}\right\|$. The unit quaternion $\bar{q_l}$ can be transformed from $\mathbf{R}\left(\bar{q}_{l}\right)=\left[\mathbf{n}_{\mathbf{e}} \ \mathbf{v}_{\mathbf{e}} \ \lfloor\mathbf{n}_{\mathbf{e}}\times\rfloor \mathbf{v}_{\mathbf{e}} \right]$, where $\mathbf{n}_{\mathbf{e}}$ and $\mathbf{v}_{\mathbf{e}}$ are the unit 3D vectors of $\mathbf{n}_{l}$ and $\mathbf{v}_{l}$.

Moreover, for each robot $i$, we adopt the simple projective line measurement model \cite{Bartoli} to describe the 2D line distance from two line endpoints, $\mathbf{x}_{s_i} = \left[u_{s_i} \ v_{s_i} \ 1\right]^{\top}$ and $\mathbf{x}_{e_i} = \left[u_{e_i} \ v_{e_i} \ 1\right]^{\top}$ to the 2D line segment:
\begin{equation}
	\begin{aligned}
		{ }^{C_i}\mathbf{z}_{l}=\left[\begin{array}{ll}
			\frac{\mathbf{x}_{s_i}^{\top} \mathbf{l}_i}{\sqrt{l_{1}^{2}+l_{2}^{2}}} & \frac{\mathbf{x}_{e_i}^{\top} \mathbf{l}_i}{\sqrt{l_{1}^{2}+l_{2}^{2}}}
		\end{array}\right]^{\top},
        \label{eq:ln3}
    \end{aligned}
\end{equation}
where $\mathbf{l}_i = \left[l_1 \ l_2 \ l_3\right]^{\top}$ denotes the 2D line representation. The line measurement can be projected from the 3D line in the camera frame as in \cite{Yang2}:
\begin{gather}
	\left[\begin{array}{c}
		l_1 \\ l_2 \\ l_3 \end{array}\right]=
	\left[\begin{array}{cccccc}
		f_{v_i} & 0 & 0 & 0 & 0 & 0\\
		0 & f_{u_i} & 0 & 0 & 0 & 0\\
		-f_{v_i} c_{u_i} & -f_{u_i} c_{v_i} & f_{u_i} f_{v_i} & 0 & 0 & 0\\
	\end{array}\right] { }^{C_i} \mathbf{L},
	\label{eq:ln4}
\end{gather}
where $f_{u_i}$, $f_{v_i}$, $c_{u_i}$, $c_{v_i}$ are the camera intrinsic parameters, and ${ }^{C_i} \mathbf{L} = \left[{ }^{C_i}d_l { }^{C_i} \mathbf{n}_{e}^\top \quad { }^{C_i} \mathbf{v}_{e}^\top\right]^\top$ is the Plücker coordinate representation of the 3D line in the camera frame. The line transformation from the global frame to the camera frame can be written as:
\begin{gather}
	\nonumber
	{ }^{C_i} \mathbf{L}=
	\left[\begin{array}{cc}
		{ }_{I_i}^{C_i}\mathbf{R} & \lfloor { }^{C_i}\mathbf{P}_{I_i}\times\rfloor { }_{I_i}^{C_i}\mathbf{R}\\
		\mathbf{0}_3 & { }_{I_i}^{C_i}\mathbf{R}\\
	\end{array}\right]
	{ }^{I_i} \mathbf{L}
\end{gather}
and
\begin{gather}
	{ }^{I_i} \mathbf{L}=
	\left[\begin{array}{cc}
		{ }_{G}^{I_i}\mathbf{R}\left(\bar t_i \right) & -{ }_{G}^{I_i}\mathbf{R}\left(\bar t_i \right) \lfloor { }^{G}\mathbf{P}_{I_i}\left(\bar t_i \right)\times\rfloor\\
		\mathbf{0}_3 & { }_{G}^{I_i}\mathbf{R}\left(\bar t_i \right)\\
	\end{array}\right]
	{ }^{G} \mathbf{L},
	\label{eq:ln5}
\end{gather}
where ${ }^{I_i} \mathbf{L}$ and ${ }^{G} \mathbf{L}$ are the Plücker line representations in the IMU frame and the global frame, respectively.

\subsection{Independent Point and Line Feature Update}
\label{section:ind}
To perform the independent point or line feature update, a standard MSCKF update \cite{Mourikis} will be applied to each robot. In particular, we collect all of the point and line measurements over the current sliding window. By stacking the measurements of one point or line, we can triangulate the point feature or line feature utilizing the estimate of the IMU poses. To simplify the notation, let $\tilde{\mathbf{x}}_{f}$ denotes either a point feature or a line feature, and the measurement residual of robot $i$ can be linearized as:
\begin{align}
	\mathbf{r}_i =  \mathbf{h}\left(\tilde{\mathbf{x}}_i, {}^{G}  \tilde{\mathbf{x}}_{f}\right)+\mathbf{w}_{i} 
	\simeq \mathbf{H}_{i, x}\tilde{\mathbf{x}}_{i}+\mathbf{H}_{i, f}{ }^{G}\tilde{\mathbf{x}}_{f}+\mathbf{w}_{i},
	\label{eq:update1}
\end{align}
where $\mathbf{r}_{i}$ is the residual of a point or line measurement. $\mathbf{H}_{i, x}$ and $\mathbf{H}_{i, f}$ denote the Jacobians w.r.t. the state vector and the feature, respectively. $\mathbf{w}_{i}$ denotes the noise vector corresponding to the point or line feature.

After that, we perform the left nullspace projection by applying the QR decompositon to $\mathbf{H}_{i, f}$ in (\ref{eq:update1}) as:
\begin{align}
	\left[\begin{array}{c}
		\mathbf{r}_{i}^1 \\
		\mathbf{r}_{i}^2
	\end{array}\right]=\left[\begin{array}{c}
		\mathbf{H}_{i, x}^1 \\
		\mathbf{H}_{i, x}^2
	\end{array}\right] \tilde{\mathbf{x}}_{i}+\left[\begin{array}{c}
		\mathbf{H}_{i, f}^1 \\
		\mathbf{0}
	\end{array}\right]{ }^{G} \tilde{\mathbf{x}}_{f}+\left[\begin{array}{c}
		\mathbf{w}_{i}^1 \\
		\mathbf{w}_{i}^2
	\end{array}\right].
	\label{eq:update2}
\end{align}
In this expression, $\mathbf{r}_{i}^2$ is only related to the state vector $\tilde{\mathbf{x}}_{i}$. Hence robot $i$ will perform an EKF update using $\mathbf{r}_{i}^2$, while $\mathbf{r}_{i}^1$ will be dropped.

\subsection{Common Point and Line Feature Update}
Note that neighboring robots might observe a common point or line feature. Hence, we will further exploit both point and line feature constraints among neighbors to improve the localization accuracy. The robots can communicate with their neighbors to share information. 

Robot $i$ and its neighbors will apply the linearization (\ref{eq:update1}) and the left nullspace projection (\ref{eq:update2}) to the common feature, denoted as ${ }^{G} \tilde{\mathbf{x}}_{f}$. As in Sec \ref{section:ind}, robot $i$ will use $\mathbf{r}_{i}^2$ for an EKF update. However, instead of dropping $\mathbf{r}_{i}^1$, robot $i$ will exploit shared information from its neighbors. It will construct a new residual system that depends on the common point or line feature ${ }^{G} \tilde{\mathbf{x}}_{f}$ by stacking the top parts in (\ref{eq:update2}) associated with itself and its neighbors as in \cite{Zhu}:
\begin{align}
	\nonumber
	\left[\begin{array}{c}
		\mathbf{r}_{i}^1 \\
		\mathbf{r}_{i_1}^1 \\
		\vdots \\
		\mathbf{r}_{i_j}^1
	\end{array}\right]=diag\left(\left[\begin{array}{c}
		\mathbf{H}_{i, x}^1 \\
		\mathbf{H}_{i_1, x}^1 \\
		\vdots \\
		\mathbf{H}_{i_j, x}^1
	\end{array}\right]\right) 
    \left[\begin{array}{c}
    	\tilde{\mathbf{x}}_{i} \\
    	\tilde{\mathbf{x}}_{i_1} \\
    	\vdots \\
    	\tilde{\mathbf{x}}_{i_j}
    \end{array}\right]+ \\
    \left[\begin{array}{c}
		\mathbf{H}_{i, f}^1 \\
		\mathbf{H}_{i_1, f}^1 \\
		\vdots \\
		\mathbf{H}_{i_j, f}^1 \\
	\end{array}\right]{ }^{G} \tilde{\mathbf{x}}_{f}+\left[\begin{array}{c}
		\mathbf{w}_{i}^1 \\
		\mathbf{w}_{i_1}^1 \\
		\vdots \\
		\mathbf{w}_{i_j}^1 \\
	\end{array}\right],
    \label{eq:update3}
\end{align}
where $diag$ denotes the block-diagonal matrix, and $i_1 \ldots i_j$ denote the neighbors of robot $i$. Then, we utilize the left nullspace projection to the stacked common point or line feature Jacobian matrix in (\ref{eq:update3}) and obtain a new residual system that is independent of the common feature as:
\begin{align}
	\mathbf{r}_{i}^\prime =\left[\begin{array}{cccc}
		\mathbf{H}_{i, x}^{\prime} \quad \mathbf{H}_{i_1, x}^{\prime} \quad \cdots \quad \mathbf{H}_{i_j, x}^{\prime}
	\end{array}\right]
	\left[\begin{array}{c}
		\tilde{\mathbf{x}}_{i} \\
		\tilde{\mathbf{x}}_{i_1} \\
		\vdots \\
		\tilde{\mathbf{x}}_{i_j}
	\end{array}\right]+	\mathbf{w}_{i}^\prime.
    \label{eq:update4}
\end{align}

In order to guarantee the consistency of estimation, we adopt the CI-EKF algorithm in \cite{Zhu}, where the weights of the CI are $\omega_i > 0, \ \omega_{i_l} > 0$, and $\omega_i + \displaystyle\sum\limits_{l=1}^{j} \omega_{i_l} = 1$. The Kalman gain of robot $i$ is given by:
\begin{align}
	\mathbf{K}_i = \frac{ \mathbf{P}_{i, k+1 \mid k} \mathbf{H}_{i, x}^{\prime \top}}{\omega_{i}} \left(\sum_{r \in \mathcal{N}_{i}} \frac{1}{\omega_{r}} \mathbf{H}_{r, x}^{\prime} \mathbf{P}_{r, k+1 \mid k} \mathbf{H}_{r, x}^{\prime \top}+\mathbf{R}_{i} \right)^{-1},
	\label{eq:update5}
\end{align}
where $\mathcal{N}_{i}$ denotes the set of robot $i$'s neighboring robots that the current common feature can be tracked, and $\mathbf{R}_{i}$ denote the covariance matrix associated with $\mathbf{w}_{i}^\prime$. Then, the state correction of robot $i$ can be written as:
\begin{align}
	\Delta \mathbf{x}_{i, k}=\mathbf{K}_i \mathbf{r}_{i}^{\prime}.
	\label{eq:update6}
\end{align}
The state covariance matrix of robot $i$ is updated using the CI as:
\begin{align}
	\mathbf{P}_{i, k+1 \mid k+1}=\frac{1}{\omega_{i}} \left(\mathbf{I} - \mathbf{K}_i \mathbf{H}_{i, x}^{\prime}\right) \mathbf{P}_{i, k+1 \mid k}.
	\label{eq:update7}
\end{align}

\section{SIMULATIONS AND EXPERIMENTS}
In this section, we utilize Monte-Carlo simulations and real-world datasets to verify that common line features can improve localization accuracy in cooperative cases, and line features can also improve the accuracy in independent cases. We compare our PL-CVIO algorithm with the previous works in Table \ref{table:compare} under two different environments, where \textit{low-feature} scenes contain a few features and \textit{rich-feature} scenes contain enough features. As shown in Table.\ \ref{table:compare}, P-VIO denotes the independent MSCKF algorithm \cite{Mourikis}, PL-VIO denotes the independent point-line MSCKF algorithm \cite{Yang1}, P-CVIO denotes our previous work CVIO \cite{Zhu}, IPL-CP-CVIO denotes the algorithm which not only utilizes independent point-line features from each robot but also collects the common point features from the neighbors, and PL-CVIO uses both independent and common point-line features as in this paper.

\begin{table}[h]
	\centering
	\caption{Descriptions of various algorithms to be compared in the simulations and experiments.}
	\renewcommand{\arraystretch}{1.5} 
	\begin{tabular}{ c c c c }
		\toprule
		\textbf{Algorithm} & \textbf{Independent Features} & \textbf{Common Features}\\
		\midrule
		P-VIO \cite{Mourikis} & Points & \xmark \\
		PL-VIO \cite{Yang1} & Points w/ Lines & \xmark \\
		P-CVIO \cite{Zhu} & Points & Points \\
		IPL-CP-CVIO & Points w/ Lines & Points \\
		PL-CVIO & Points w/ Lines & Points w/ Lines \\
		\bottomrule
	\end{tabular}
	\label{table:compare}
\end{table}

\begin{table*}[t]
	\centering
	\caption{The RMSE of the orientation / position (degrees / meters) of three robots using three different algorithms in different EuRoC datasets. \textbf{R} denotes the rich-feature environments with enough point-line features, and \textbf{L} denotes the low-feature cases. The average denotes mean of all three rooms per algorithm per robot per environment (rich-feature/low-feature). R0, R1, and R2 represent three robots following three different trajectories in each environment.}
	\renewcommand{\arraystretch}{1.5} 
	\begin{tabular}{ c | c c c c | c c c c }
		\toprule
		& \textbf{V1\_01\_R} & \textbf{V1\_02\_R} & \textbf{V1\_03\_R} & \textbf{Average} & \textbf{V1\_01\_L} & \textbf{V1\_02\_L} & \textbf{V1\_03\_L} & \textbf{Average}\\
		\midrule
		R0 P-VIO & 0.481 / 0.260 & 0.621 / 0.064  & 0.874 / 0.061  & 0.659 / 0.128 &1.277 / 0.483  & 0.717 / 0.177  & 1.184 / 0.684 & 1.060 / 0.448 \\
		R0 P-CVIO &  0.091 / 0.056   & 0.157 / 0.022   & 0.118 / 0.027  & 0.122 / 0.035 &  0.524 / 0.148  & 0.449 / 0.080  & 0.743 / 0.299 & 0.572 / 0.176 \\
		R0 PL-CVIO & 0.090 / 0.047 &  0.147 / 0.021& 0.101 / 0.025 & 0.113 / 0.031 & 0.159 / 0.078 & 0.167 / 0.064 & 0.182 / 0.099 & 0.169 / 0.080\\
		\midrule
		R1 P-VIO & 1.166 / 0.205 & 0.167 / 0.049  & 0.419 / 0.049  & 0.584 / 0.101 &  0.888 / 0.152  & 0.785 / 0.170  & 0.775 / 0.150 & 0.816 / 0.157 \\
		R1 P-CVIO & 0.104 / 0.060 & 0.183 / 0.026   & 0.127 / 0.026  &  0.138 / 0.037 &  0.584 / 0.137  & 0.613 / 0.130  & 0.733 / 0.075 & 0.643 / 0.144 \\
		R1 PL-CVIO & 0.089 / 0.052 &  0.176 / 0.021 &  0.096 / 0.026 & 0.120 / 0.033 & 0.213 / 0.092 & 0.231 / 0.078  & 0.249 / 0.074 & 0.231 / 0.081\\
		\midrule
		R2 P-VIO & 0.960 / 0.132 & 0.230 / 0.078  & 0.325 / 0.062  & 0.505 / 0.091 & 1.589 / 0.596 & 1.493 / 0.195  & 0.676 / 0.202 & 1.253 / 0.331\\
		R2 P-CVIO & 0.099 / 0.056 & 0.170 / 0.023   & 0.123 / 0.025  & 0.131 / 0.035 &  0.603 / 0.161 & 0.690 / 0.179  & 0.538 / 0.165 & 0.610 / 0.168\\
		R2 PL-CVIO & 0.095 / 0.056 &  0.167 / 0.022 &  0.109 / 0.021 & 0.127 / 0.033 &  0.150 / 0.096 &  0.167 / 0.081 & 0.182 / 0.073 & 0.166 / 0.083\\
		\bottomrule
	\end{tabular}
	\label{table:result}
\end{table*}

\subsection{Monte-Carlo Simulations}
For our Monte-Carlo simulations, we utilize a group of three robots. \textit{Robot 0} in the group follows the real trajectory of a dataset, and the trajectories of \textit{robot 1} and \textit{robot 2} are created by adding position and orientation offsets to the real one. After that, the 3D features and the corresponding 2D measurements are generated if the number of the point or line measurements is below the threshold in the current frame. Then, the constraints of the same feature from one robot and the constraints of the common features from neighbors are collected and utilized to update the current state.

\begin{figure}[t]
	\begin{center}
		\includegraphics[scale=0.37]{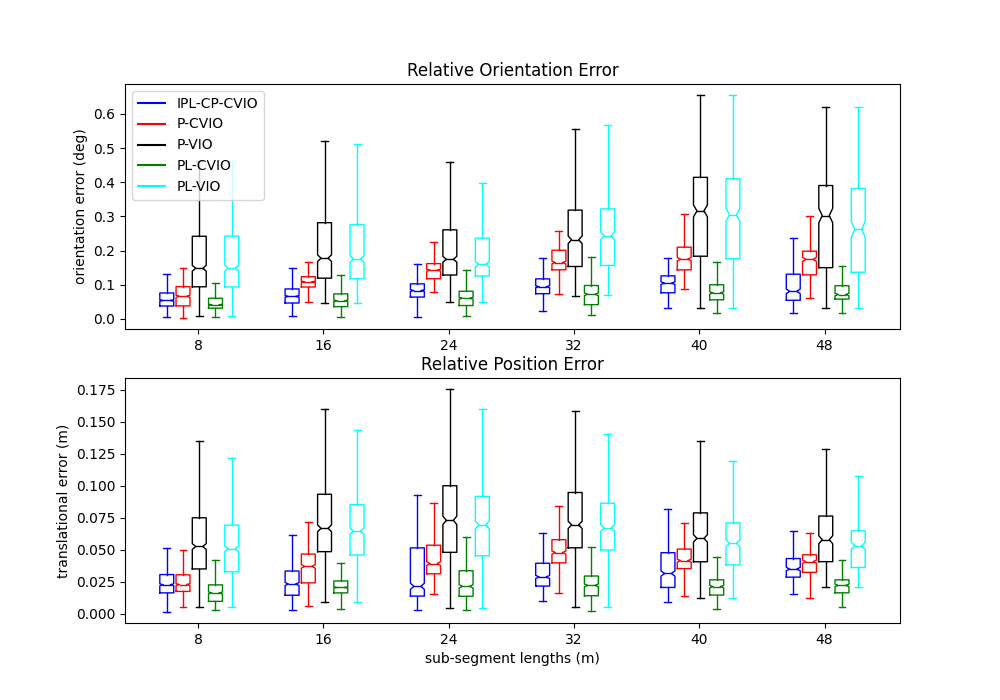}
	\end{center}
	\caption{Boxplot of the statistics of the Monte-Carlo simulation under the rich-feature Udel\_gore environment by extracing 150 points per frame, and 50 lines if the line update is used.}
	\label{fig:sim1}
\end{figure} 
\begin{figure}[t]
	\begin{center}
		\includegraphics[scale=0.37]{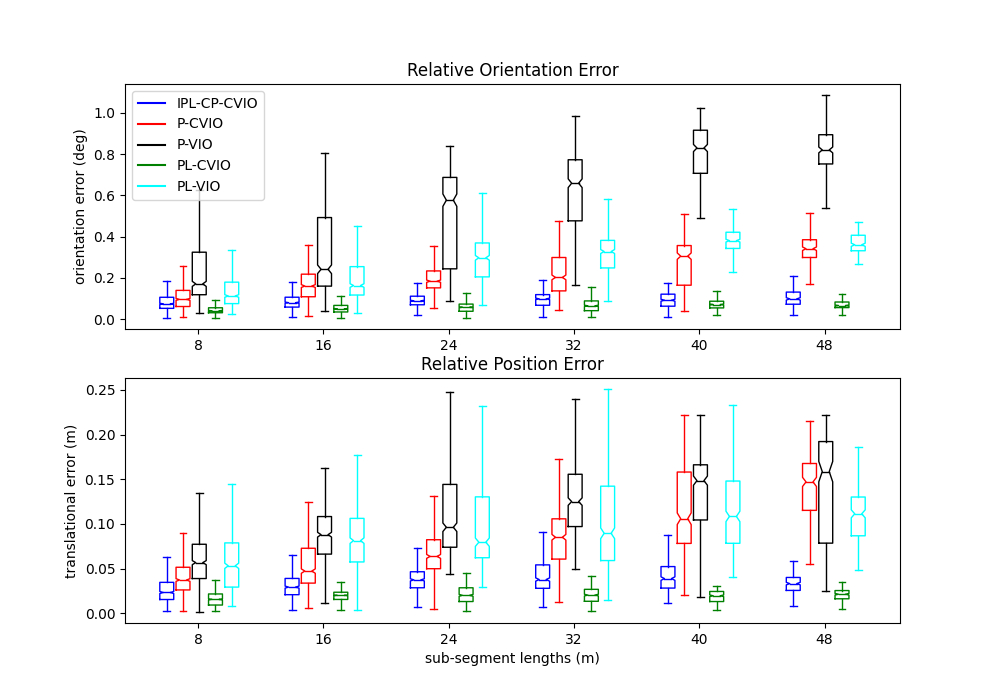}
	\end{center}
	\caption{Boxplot of the statistics of the Monte-Carlo simulation under low-feature Udel\_gore environment by extracing 50 points per frame, and 50 lines if the line update is used.}
	\label{fig:sim2}
\end{figure}

The low-feature and rich-feature environments are divided by extracting different numbers of point features. In the rich-feature environments, the number of point features is 150 and the number of line features is 50 in each frame. We reduce the number of point features to 50 for the low-feature cases. For both of these two environments, we utilize the First-Estimation Jacobian (FEJ) and online camera-IMU calibration \cite{Geneva}. After running 30 Monte-Carlo loops, the statistics of the relative orientation error (ROE) and the relative position error (RPE) under the rich-feature or low-feature Udel\_gore dataset are shown in Fig.\ \ref{fig:sim1} and Fig.\ \ref{fig:sim2}, respectively. We can see that our PL-CVIO algorithm outperforms all other algorithms in both environments. Especially in the low-feature case, we can find out that the common line can reduce the ROE and RPE obviously (blue and red bar) as in Fig.\ \ref{fig:sim2}. Moreover, an interesting discovery is that PL-VIO outperforms P-CVIO if a limited number of points are observed in each frame. In this case, the number of common point features is also limited, and hence the cooperative method P-CVIO that relies on only common point features has limited resources to resort to. In contrast, the methods PL-VIO and PL-CVIO that further exploit line features exhibit better performance while PL-CVIO achieves the best performance as it exploits not only point and line features but also cooperation with neighbors. 

Additionally, we simulate our PL-CVIO algorithm in all of the EuRoC V1 datasets \cite{Burri} and compare it with P-VIO and P-CVIO in both low-feature and rich-feature environments. The RMSE of the orientation and position of each robot and the mean RMSE of each algorithm in each environment are recorded in Table \ref{table:result}. The RMSE results show that our PL-CVIO algorithm outperforms P-CVIO and P-VIO in all simulated scenarios. Especially in low-feature environments, the PL-CVIO improves the RMSE of orientation and position dramatically.

\subsection{Experiments}
For the real-world experiments, the position and orientation of each robot are initialized corresponding to the ground truth. The point features are extracted from each frame using FAST \cite{Rosten}, and are tracked crossing frames or matched with the point observations from other robot utilizing ORB \cite{Rublee} with an 8-point RANSAC algorithm. At the same time, line segments are extracted by leveraging the LSD \cite{Von} and tracked by LBD \cite{Zhang}. Besides, we add some outlier elimination strategies to remove the line segment where (1) the LBD distance is larger than 50; (2) the length of the line segment is smaller than 50 pixels; (3) the distance between the origin and this line is smaller than 0.1 or larger than 100; (4) the line disparity is too small to avoid singularity when applying the SVD \cite{Bradski} to triangulate this line.
\begin{figure}[t]
	\begin{center}
		\includegraphics[scale=0.37]{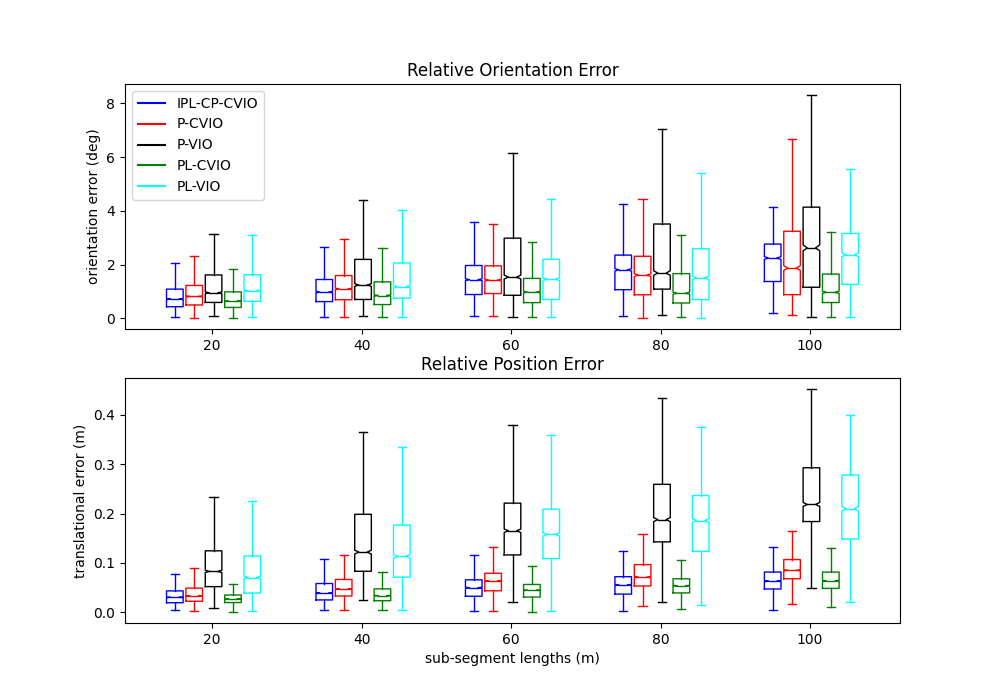}
	\end{center}
	\caption{Boxplot of the result of \textit{Robot 0} (Room 1) in the TUM Visual-Inertial Dataset by extracing 200 points per frame, and 50 lines if the line update is used.}
	\label{fig:exp1}
\end{figure}

\begin{figure}[t]
	\begin{center}
		\includegraphics[scale=0.37]{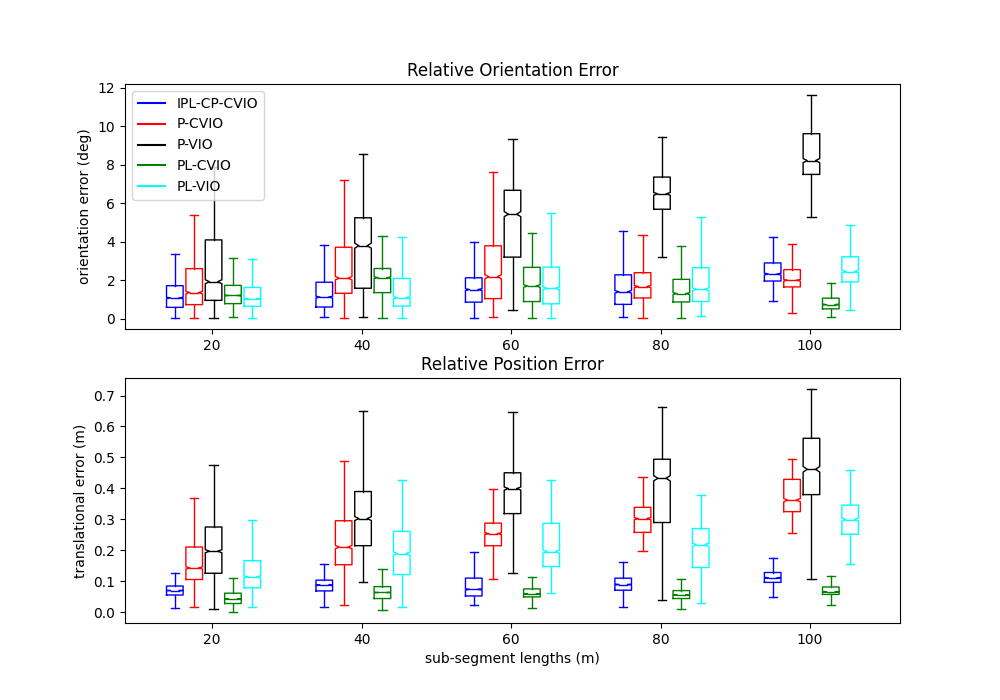}
	\end{center}
	\caption{Boxplot of the result of \textit{Robot 0} (Room 1) in the TUM Visual-Inertial Dataset by extracing 50 points per frame, and 50 lines if the line update is used.}
	\label{fig:exp2}
\end{figure}

We evaluate our PL-CVIO algorithm in the TUM Visual-Inertial Dataset Rooms 1, 3, and 5 \cite{Sturm}, where the IMU frequency is 200 Hz and the camera frequency is 20Hz. We load all three datasets of the same room and run all five algorithms with three robots separately. Besides, we extract a different number of point features to imitate low-feature and rich-feature environments. As a result, we show the experimental results of our PL-CVIO algorithm compared with the other four algorithms in respectively rich-feature environments as in Fig.\ \ref{fig:exp1} and low-feature environments as in Fig.\ \ref{fig:exp2}. We also show the RMSE of the orientation and position of each robot by utilizing different algorithms in the TUM dataset as in Table \ref{table:result2}. From the ROE/RPE and the RMSE results, it is clear that line features can improve the accuracy of P-VIO and the common point-line features can improve the performance of the P-CVIO. Besides, the lines improve the performance obviously in the low-feature scenes by comparing the P-VIO and PL-VIO, as well as P-CVIO and IPL-CP-CVIO in Table.\ \ref{table:result2}. Additionally, our PL-CVIO outperforms all four algorithms in all experiment cases.

\begin{table}[t]
	\centering
	\caption{The RMSE of the orientation / position (degrees / meters) of three robots under the low-feature environments by using five different algorithms in the TUM Visual-Inertial dataset.}
	\renewcommand{\arraystretch}{1.5} 
	\begin{tabular}{ c c c c }
		\toprule
		\textbf{Algorithm} & \textbf{Robot 0} & \textbf{Robot 1} & \textbf{Robot 2} \\
		\midrule
		P-VIO & 7.473 / 0.442 & 4.357 / 0.313  & 5.468 / 0.372  \\
		PL-VIO & 2.367 / 0.295 & 1.798 / 0.254  & 1.872 / 0.240 \\
		P-CVIO & 2.301 / 0.377   & 3.824 / 0.267   &  2.616 / 0.263  \\
		IPL-CP-CVIO & 1.905 / 0.098   & 1.722 / 0.115   & 1.542 / 0.095  \\
		PL-CVIO & 1.349 / 0.061 & 1.665 / 0.086  & 1.379 / 0.067 \\
		\bottomrule
	\end{tabular}
	\label{table:result2}
\end{table}

\section{CONCLUSIONS}
In this paper, we have proposed a fully distributed point-line cooperative visual-inertial navigation system. We compared the performance of the proposed algorithm with four other algorithms under rich-feature or low-feature environments in both Monte-Carlo simulations and real-world datasets. All of the results indicated that our PL-CVIO outperformed the independent MSCKF and CVIO. Also, we verified that the line feature can improve the accuracy of localization in independent cases, and the common line features can perform better in cooperative cases.


\end{document}